%% file: main.tex
\documentclass[sigconf]{acmart}

\usepackage{booktabs} 
\usepackage{algorithm}
\usepackage{algorithmic}
\setcopyright{rightsretained}

\begin{document}
\title{Supervised Typing of Big Graphs using Semantic Embeddings}

\author{Mayank Kejriwal}
\affiliation{%
  \institution{Information Sciences Institute \\ USC Viterbi School of Engineering}
  \streetaddress{4676 Admiralty Way, Ste. 1001}
  \city{Marina Del Rey} 
  \state{California} 
  \postcode{90292}
}
\email{kejriwal@isi.edu}

\author{Pedro Szekely}
\affiliation{%
  \institution{Information Sciences Institute \\ USC Viterbi School of Engineering}
  \streetaddress{4676 Admiralty Way, Ste. 1001}
  \city{Marina Del Rey} 
  \state{California} 
  \postcode{90292}
}
\email{pszekely@isi.edu}

\renewcommand{\shortauthors}{Kejriwal and Szekely}

\begin{abstract}
We propose a supervised algorithm for generating type embeddings in the same semantic vector space as a given set of entity embeddings. The algorithm is agnostic to the derivation of the underlying entity embeddings. It does not require any manual feature engineering, generalizes well to hundreds of types and achieves near-linear scaling on Big Graphs containing many millions of triples and instances by virtue of an incremental execution. We demonstrate the utility of the embeddings on a type recommendation task, outperforming a non-parametric feature-agnostic baseline while achieving 15x speedup and near-constant memory usage on a full partition of DBpedia. Using state-of-the-art visualization, we illustrate the agreement of our extensionally derived DBpedia type embeddings with the manually curated domain ontology. Finally, we use the embeddings to probabilistically cluster about 4 million DBpedia instances into 415 types in the DBpedia ontology.       
\end{abstract}

\copyrightyear{2017} 
\acmYear{2017} 
\setcopyright{acmcopyright}
\acmConference{SBD'17}{May 19, 2017}{Chicago, IL, USA}\acmPrice{15.00}\acmDOI{http://dx.doi.org/10.1145/3066911.3066918}
\acmISBN{978-1-4503-4987-1/17/05}
%
%







\maketitle

\input{samplebody-conf}

\bibliographystyle{ACM-Reference-Format}
\bibliography{sigproc} 

\end{document}

%% file: samplebody-conf.tex
In recent years, the \emph{distributional semantics} paradigm has been used to great effect in natural language processing (NLP) for embedding words in vector spaces \cite{turian2010}. The distributional hypothesis (also known as \emph{Firth's axiom}) states that the meaning of a word is determined by its \emph{context} \cite{distributionalhypothesis}. Algorithms like word2vec use neural networks on large corpora of text to embed words in \emph{semantic} vector spaces such that contextually similar words are close to each other in the vector space \cite{word2vec}. Simple arithmetic operations on such embeddings have yielded semantically consistent results (e.g. $King - Man + Woman = Queen$). 
\begin{figure}[t]
\centering
\includegraphics[height=4.3cm, width=7.5cm]{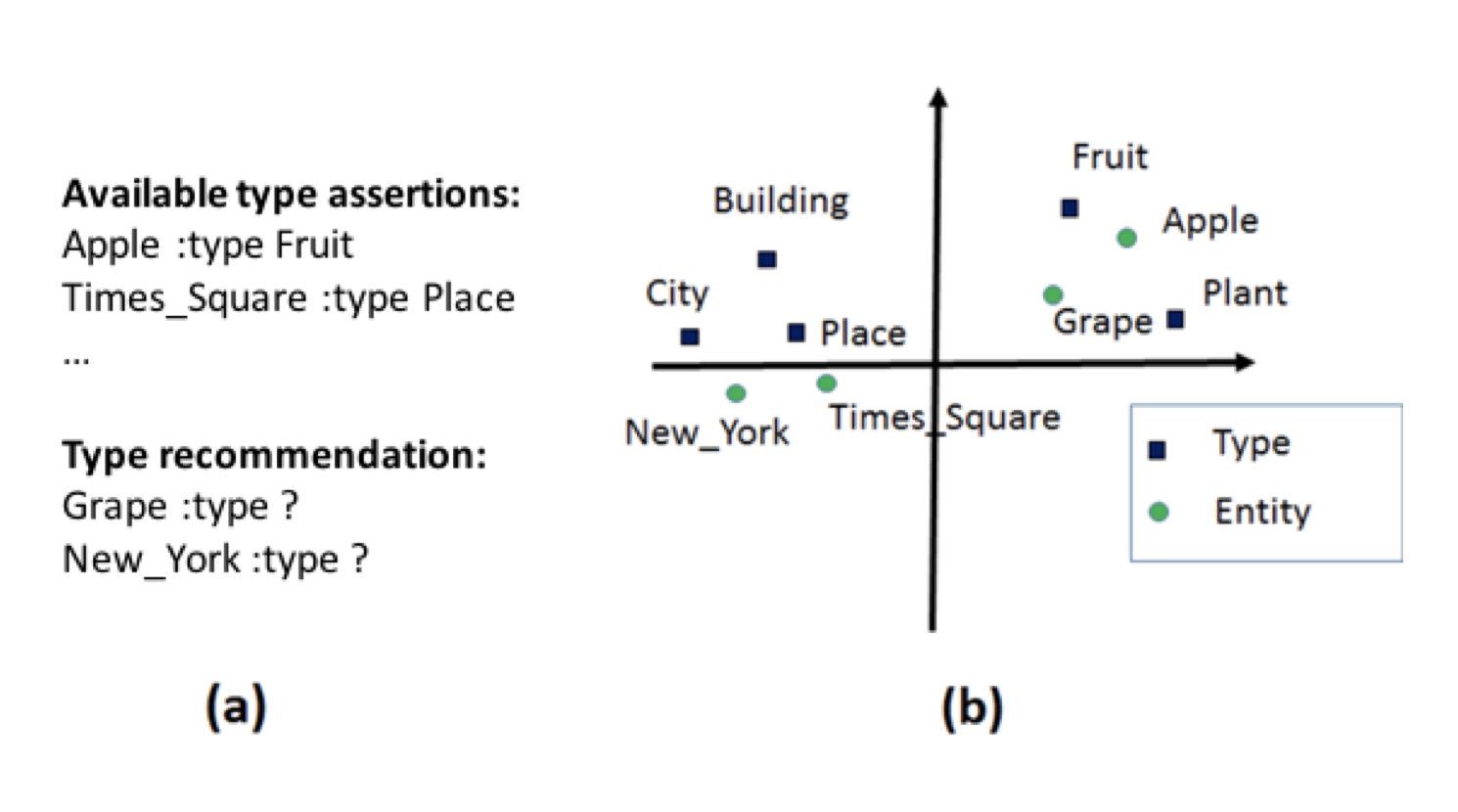}
\caption{Visual intuition behind semantic embeddings and how such embeddings can be used for the type recommendation task in a supervised setting}
\label{example} 
\end{figure}

Recent work has extended such neural embedding techniques, traditionally introduced only for natural language word sequences, to alternate kinds of data, including entities in large knowledge graphs like DBpedia \cite{graph2vec,rdf2vec}. The basic approach is to convert an instance-rich knowledge graph into sets of sequences of graph nodes by performing random walks or using graph kernels \cite{graphkernels}. NLP algorithms like word2vec are applied on the sequences to embed entities, just like words in natural language sentences \cite{rdf2vec}. 

In the Semantic Web, the domain of discourse is typically expressed by a manually curated ontology. A basic element of an ontology is a \emph{type}, also called a \emph{class}. \emph{Type assertion} statements relate entities (i.e. instances) in a knowledge base (KB) to the domain ontology, which can then be used to \emph{infer} more facts about entities.

Given the crucial role played by types in mediating between domain ontologies and instance-rich KBs, a natural question is whether types can be embedded in the same semantic vector space as entities, and whether data-driven type embeddings can be used to reason about, and visualize, ontologies. For example, one could use these embeddings to ask whether the data adequately capture ontological semantics (Section \ref{exp2}), and to recommend types for new entities in the knowledge base (Section \ref{exp1}). Unfortunately, type embeddings are difficult to directly derive from graphs because big knowledge graphs are sparse in type assertion statements compared to the number of unique instances and facts. In DBpedia, for example there are over 17 million (non-type) triples, and almost 4 million unique entities; the number of type assertion statements is also about 4 million, meaning that there is usually only one type assertion per entity. In many cases, the type assertions can be trivial (e.g. owl\#Thing). Another problem is that types are typically asserted as objects, not subjects, in the KB; hence, a random walk cannot be initiated from a type node.

We propose a scalable proof-of-concept solution to the problem of deriving type embeddings \emph{from} entity embeddings in big graphs. A visual intuition behind the method is provided in Figure \ref{example}. Given a set of pre-generated entity embeddings, and a sparse collection of type assertion triples, we are able to robustly generate embeddings for a set of types (e.g. \emph{Fruit}, \emph{Building}). We use these embeddings for a variety of tasks, most notably probabilistic type recommendation (e.g. recommending types such as \emph{Fruit} and \emph{Plant} for a new entity like \emph{Grape}), intensional semantics visualization, and probabilistic type clustering over large graphs (Section \ref{evaluations}). Specific empirical highlights are noted below. To the best of our knowledge, this is the first work that presents a \emph{feature-agnostic} supervised typing of Big Graphs.

{\bf Empirical Highlights:} Preliminary empirical results on a partition of DBpedia show that our algorithm achieved run-time speedups by more than a factor of 15 on the type recommendation task compared to non-parametric nearest-neighbors baselines, with superior recall on two relevance criteria.  Visualizations of the type embeddings using the DBpedia ontology as a ground-truth show that they are in good agreement with the intensional semantics expressed by the ontology. The scalability of the model enabled us to probabilistically cluster almost 4 million DBpedia instances into 415 types on a serial machine in under 50 hours.

\section{Related Work}\label{relatedwork}
Semantic vector space embeddings have witnessed much research in recent years, with neural word embedding algorithms (e.g. word2vec \cite{word2vec} and glove \cite{glove}) achieving state-of-the-art performance on a number of NLP tasks (e.g. dependency parsing) \cite{chen2014}. The success of word embedding approaches has led to a renewed interest in graph-based communities for embedding graphs.  A famous example is DeepWalk, which applies word embedding techniques on random walk sequences on a graph to embed nodes in the graph to vectors \cite{deepwalk}. In the Semantic Web, variants of this strategy were recently applied to DBpedia and Wikidata, and the embedded entities were used in several important problems, include content-based recommendation and node classification \cite{rdf2vec},\cite{recommender}. 
Some other influential examples of such \emph{knowledge graph embeddings} (KGEs), which is an active area of research, include (but are not limited to) \cite{kge1}, \cite{kge2}, \cite{kge4}, \cite{kge3}. An important aspect of this research is automatic knowledge base construction and completion (AKBC), to which this work is related \cite{akbc}, \cite{akbc2}. A major difference is that, because of an additional layer of semantic abstraction (types vs. entities), we can afford to infer types without incrementally  training the model such as in \cite{psdvec} or any other details of how the entity embeddings were derived. We also do not rely on natural language context of any kind \cite{dbpediaETNLP}.  

In this paper, we do not seek to develop a \emph{new} learning algorithm for graph (including knowledge graph) or word embeddings; instead, the goal is to use an existing publicly available set of graph entity embeddings to extensionally model types. To the best of our knowledge, this is the first paper to derive the embedding of \emph{schema-level} elements (like types) directly using the embeddings of \emph{instance-level} elements like entities.  Because our method does not make underlying assumptions about the entity embeddings, it is general and can be applied to any set of entity embeddings.

The type \emph{recommendation} problem to which we apply the type models is closely connected to the type \emph{prediction} problem studied in prior work, a good example being \emph{Typifier} \cite{typifier}. Unlike Typifier, which is not embedding-based and relies on manually devised features (e.g. data and pseudo-schema features \cite{typifier}), our approach is feature-agnostic. Other examples of feature-centric type recommenders are the systems in \cite{typer1}, \cite{typer2}. Due to the difficulty in automating feature construction, feature-agnostic systems are still quite rare; for example, in the Semantic Web, only a recent work achieved competitive performance at scale for feature-agnostic node classification \cite{rdf2vec}. Furthermore, because embeddings are a very general framework, we use the embeddings not just for type embeddings but also visualization and online clustering, which cannot be handled by the other (special-purpose) type recommenders.

\section{Approach}\label{approach}
\subsection{Framework}
We lay the groundwork for this section by formally defining a \emph{typed} knowledge base (t-KB) and related terminology  below:

{\bf Definition.} Given a set $\mathcal{I}$ of Internationalized Resource Identifiers (IRIs), a set $\mathcal{B}$ of blank nodes and a set $\mathcal{L}$ of literals, a typed RDF Knowledge Base $\mathcal{T}$ is a set of RDF triples (i.e. $\subseteq \{\mathcal{I} \cup \mathcal{B}\} \times \mathcal{I} \times \{\mathcal{I} \cup \mathcal{B} \cup \mathcal{L}\}$) such that $\forall (s,p,o) \in \mathcal{T} \rightarrow \exists t \in \mathcal{I}, (s,:type,t) \in \mathcal{T}$, where $:type \in \mathcal{I}$ is a special type property (e.g. \emph{rdf:type}).

We denote $(s,:type,t)$ as a \emph{type assertion} statement, an arbitrary element $s$ from the \emph{entity set} $S=\{s | (s, p, o) \in \mathcal{T}\}$ as an \emph{entity}, and the set $T(s)=\{t | (s, :type, t) \in \mathcal{T}\}$ as its set of asserted types\footnote{At present, we take an extensional (or instance-driven) view of a type by identifying it by its referents (the set of explicitly declared instances) of an entity $s$. We investigate an empirical relaxation of this condition in Section \ref{evaluations}.}. Similar to the entity set $S$, we denote $T=\bigcup_{s \in S}T(s)$ as the \emph{type set} of knowledge base $\mathcal{T}$. Finally, we denote a \emph{type-only} KB (t-KB) $\mathcal{T}'$ as the subset of the typed knowledge base $\mathcal{T}$ that contains exactly the type assertions in $\mathcal{T}$.
Although each entity $s$ is \emph{represented} by an IRI per the RDF data model, an alternate representation is as an \emph{embedding} in a real-valued $d$-dimensional space:  

{\bf Definition (entity embedding).} A $d$-dimensional \emph{entity embedding} representation of an entity $s$ is a mapping from $s$ to a real-valued vector $\overrightarrow{s}$ that is constrained to lie on the unit-radius hypersphere in $d$-dimensional space.

The constraint in the definition above ($\Sigma_i \overrightarrow{s}[i]^2 = 1$) ensures that the entity embedding is l2-normalized, which simplifies distance calculations considerably by equating cosine similarity between the two vectors with a dot product. 

Concerning the actual learning of entity embeddings, we noted in Section \ref{relatedwork} that recent work has successfully managed to learn embeddings (in spaces with only a few hundred dimensions) on large datasets like DBpedia by applying neural embedding algorithms like word2vec on graph node sequences \cite{rdf2vec}. Typically, such embeddings do not include types. A practical reason is that, in big knowledge graphs like DBpedia, the graphs $\mathcal{T}'$ and $\mathcal{T}-\mathcal{T}'$ are released as separate files (we provide links in Section \ref{evaluations}), and entity embeddings are only trained over the latter. A more serious reason, pointed out in the introduction, is sparsity of assertions and the observation that $T \cap S$ is typically empty.

To address problems of sparsity and robustness, we attempt to embed types into the same vector space as entities (thereby leveraging the enormous context of entities). Formally, we define a type embedding below.

{\bf Definition (type embedding).} Given an entity set $S$ and a type $t$, a $d$-dimensional \emph{type embedding} representation of $t$ is a mapping from $t$ to a real-valued vector $\overrightarrow{t}$ that is constrained to lie on the unit-radius hypersphere in $d$-dimensional space.

Intuitively, a `good' type embedding should have two elements: (1) be close in the vector space to entities that have that type. In Figure \ref{example} (b), for example, \emph{Fruit} is much closer to \emph{Apple} than it is to \emph{Times\_Square}; (2) be closer to `related' types than to types that are unrelated. In Figure \ref{example} (b), \emph{Building}, \emph{City} and \emph{Place} are all closer to one another than to either \emph{Fruit} or \emph{Plant}.

Clearly, the two elements above are related as they strongly rely on the data and on the context in which types are asserted or co-appear with entities that share context. In the next section, we explore how such robust embeddings can be scalably generated.

\subsection{Generating Type Embeddings}

\begin{algorithm}[t]
\begin{algorithmic}
\renewcommand{\algorithmicrequire}{\textbf{Input:}}
\renewcommand{\algorithmicensure}{\textbf{Output:}}
\REQUIRE Sets $S$ and $\vec{S}$ of entities and entity embeddings, type-only Knowledge Base $\mathcal{T}'$\\
\ENSURE Type embedding $\overrightarrow{t}$ for each type $t$ in $\mathcal{T}'$\\
\begin{enumerate}
\setlength{\itemsep}{1pt}
  \setlength{\parskip}{0pt}
  \setlength{\parsep}{0pt}
\STATE{Initialize empty dictionary $T_S$ where keys are entities and values are type-sets}
\STATE Initialize type-set $T$ of $\mathcal{T}'$ to the empty set

// First pass through $\mathcal{T}'$: collect entity-type statistics
\FORALL {triples $(s, :type, t) \in \mathcal{T}'$ such that $\overrightarrow{s} \in \vec{S}$}
\STATE Add $t$ to $T$
\STATE Add $t$ to $T_S[s]$, if it does not already exist
\ENDFOR
\STATE {\bf for all} $s \in keys(T_S)$, set $T_S[s]=|T_S[s]|$ to save memory {\bf end for}

//Second pass through $\mathcal{T}'$ to derive type embeddings
\STATE{Initialize Mean parameter dictionary $M$ such that $keys(M)=T$, and each value in $M$ is $\vec{0}$}
\FORALL {triples $(s, :type, t) \in \mathcal{T}'$ such that $s \in S$}
\STATE Update M[t] using Equation \ref{meanupdate}, using $T(s)=T_S[s]$
\ENDFOR

//Derive type embedding from $\overrightarrow{\mu_t}$
\FORALL {types $t \in keys(M)$}
\STATE Let type embedding $\overrightarrow{t}$ be the projection of $M[t]$ on $d$-dimensional hypersphere with unit radius (divide throughout by $||M[t]||_2$)
\ENDFOR
\RETURN type embeddings derived in last step
\end{enumerate}
\end{algorithmic}
\caption{Generate Type Embeddings}
\label{alg1}
\end{algorithm}

We propose a type embedding algorithm that is lightweight both in terms of run-time and memory. Algorithm \ref{alg1} provides the pseudocode for our solution. Before describing the pseudocode, we describe the intuition as follows.

Algorithm \ref{alg1} relies on two assumptions: first, a type is ultimately described by its entities. This means that, all things the same, a type should be close to as many entities having that type as possible. Second, a type should give more preference to entities that describe it exclusively. For example, suppose an entity $s_1$ has more than ten (explicitly declared) type assertions $\{t_1,t_2\ldots t_{10}\}$ while entity $s_2$ only has two type assertions $\{t_1,t_2\}$. Algorithm \ref{alg1} is set up so that $s_2$ will contribute \emph{more} to the derivation of $t_1$ and $t_2$ embeddings than $s_1$.

To operationalize these assumptions, while still being simple and scalable, Algorithm \ref{alg1} computes a \emph{weighted average} of entity embeddings to derive a type embedding. 
Specifically, in a first pass over a type-only KB $\mathcal{T}'$, the algorithm computes the number of types $|T(s)|$ asserted by entity $s$. For each new type encountered, the algorithm initializes a zero vector for the type, in the same $d$-dimensional space as the entity embeddings. Even with millions of entities, this information can be stored in memory at little cost. 

The second pass of the algorithm is \emph{incremental}. For each triple $(s, :type, t)$, we update a type vector $\overrightarrow{t}$ (initialized to $\overrightarrow{0}$) using the equation:
\begin{equation}\label{meanupdate}
\overrightarrow{t}_{new} = \overrightarrow{t}_{old} + \frac{1}{|T(s)|}\overrightarrow{s}
\end{equation}
In the notation of the algorithm, $T_S[s]=T(s)$.
Line 9 in Algorithm \ref{alg1} shows a simple way of obtaining the final type embedding $\overrightarrow{t}$ by normalizing the `final' mean vector $\overrightarrow{t}_{new}$ so that it lies on the unit-radius ($d$-dimensional) hypersphere. Normalizing ensures that the type embedding obeys the same constraints as the original entity embeddings and conforms to the type embedding definition earlier stated. A second reason for normalizing is that the computation of the cosine similarity between any vectors (whether type or entity) on the $d$-dimensional hypersphere reduces to the computation of the dot product between the vectors. The next section covers two applications of such computations.

\subsection{Applications}

{\bf Type Recommendation.} Given the type embeddings generated in the previous section, we can \emph{recommend} types (with scores) for an \emph{untyped} entity $s$ by computing the dot product between the embedding $\overrightarrow{s}$ of the entity and each of the $|T|$ type embeddings derived in Algorithm \ref{alg1}, and normalizing over the $|T|$ results (if a valid probability distribution is desired). The probability distribution can also be used to \emph{rank} a set of types with the highest-ranked type being the most \emph{relevant} suggestion for the entity. Other than `completing' knowledge bases with many untyped entities, such rankings can also assist in tasks like semantic search.

{\bf Online Clustering.} In contrast with other update-based machine learning algorithms like Stochastic Gradient Descent, we note that data can be discarded once seen, making Algorithm \ref{alg1} amenable to streaming settings. For example, we were able to probabilistically cluster the full set of DBpedia instances (yielding a $>$100 GB results file) in less than 2 days in a single-threaded computing environment.

\section{Preliminary Experiments}\label{evaluations}
The goal of this section is to illustrate the promise of the approach through some preliminary results. In Section \ref{conclusion}, we discuss future directions based on the preliminary results. 
\subsection{Preliminaries}
{\bf Datasets:} We construct five evaluation datasets by performing random stratified partitioning on the full set of DBpedia triples. We used the publicly available type-only KB\footnote{Accessed at \url{http://downloads.dbpedia.org/2015-10/core-i18n/en/instance_types_en.ttl.bz2}} for our experiments from the October 2015 release of the English-language DBpedia. This file contains all the type assertions obtained only from the mapping-based extractions, without transitive closure using the ontology. Details of the five ground-truth datasets are provided in Table \ref{datasets}. Across the full five-dataset partition, there are 3,963,983 unique instances and 415 unique types. The five datasets are roughly uniform in their representation of the overall dataset, and not subject to splitting or dataset bias, owing to random stratified partitioning.
\begin{table}[t]
\centering
\caption{Details of ground-truth datasets. The five datasets together comprise a partition of all (extensional) type assertion statements available for DBpedia.}
\begin{tabular}{|p{0.8cm}|p{1.3cm}|p{1.5cm}|p{1.3cm}|p{1.5cm}|} \hline
{\bf Data-set}&{\bf Num. triples}&{\bf Num. unique instances}&{\bf Num. unique types}&{\bf Size on disk (bytes)} \\ \hline
{D-1} & {792,835} & {792,626} & {410} & {113,015,667} \\ \hline
{D-2} & {793,500} & {793,326} & {412} & {113,124,417} \\ \hline
{D-3} & {793,268} & {793,065} & {409} & {113,104,646} \\ \hline
{D-4} & {793,720} & {793,500} & {410} & {113,168,488} \\ \hline
{D-5} & {792,865} & {792,646} & {410} & {113,031,346} \\ \hline
\end{tabular}
\label{datasets}
\end{table} 
In Section \ref{exp2}, we also compare our type embeddings to the DBpedia domain ontology\footnote{Downloaded from \url{http://downloads.dbpedia.org/2015-10/dbpedia_2015-10.nt}}.

{\bf Entity Embeddings:} The algorithms presented in this paper assume that a set of entity embeddings has already been generated. Recently, \cite{rdf2vec} publicly made available 500-dimensional embeddings for DBpedia entities by using the word2vec algorithm on graph node sequences\footnote{Accessed at \url{http://data.dws.informatik.uni-mannheim.de/rdf2vec/models/DBpedia}}. The word2vec model was trained using skip-gram, and was found to perform well on a range of node classification tasks\footnote{The authors also released Wikidata embeddings, which did not do as well on node classification and were noisier (and much larger) than the DBpedia embeddings. For this reason, we do not consider the Wikidata embeddings in this paper.}. Rather than generate our own entity embeddings (which could potentially cause bias by overfitting to the type modeling task), we used those previously generated embeddings for all experiments in this paper.

{\bf Implementation:} All experiments in this paper were run on a serial iMac with a 4 GHz Intel core i7 processor and 32 GB RAM. All code was written in the Python programming language. We used the \emph{gensim} package\footnote{\url{https://pypi.python.org/pypi/gensim}} for accessing, manipulating and computing similarity on the entity embeddings.

\subsection{Probabilistic Type Recommendation}\label{exp1}
In this experiment, we evaluate Algorithm \ref{alg1} on the \emph{probabilistic type recommendation} task. The input to the recommendation system is an entity, and the output must be a set of scored (or ranked) types that are \emph{topically relevant} to the entity. The issue of relevance, by its nature, is subjective; we present a methodology for evaluating subjective relevance shortly.

{\bf Baseline:} We employ baselines based on weighted $k$ Nearest Neighbors (kNN). The kNN algorithm is a strong baseline with some excellent theoretical properties: for example, even the 1NN algorithm is known to guarantee an error rate that is at most twice the Bayes error rate\footnote{This is the minimum possible error for a particular distribution of data.} in a given feature space and for a given distance function in the space. Because the entity embeddings are given, the kNN baseline, just like the embedding method, is \emph{feature-agnostic}; to the best of our knowledge, it is the only baseline that has this property and is not super-linear. We consider three versions with $k$ set to 1, 5 and 10 respectively. 

{\bf Baseline Complexity:} Compared to the embedding method, kNN has high time and memory complexity since it is \emph{non-parametric} and requires storing all the (training) entities in memory. In contrast, after computing the type embeddings, the embedding method has to store $|T|$ vectors, one for each type. In terms of run-time, a full pass is required over the training dataset for \emph{each} new test entity, regardless of the value of $k$. We explore the empirical consequence of this in our results.


{\bf Experiment 1: }We perform five experiments, where in each experiment, four partitions (from Table \ref{datasets}) are used as the training set, and 5000 entities are randomly sampled from the remaining partition and used as the test set. We were forced to constrain the test set to 5000 for this initial experiment because of baseline complexity. 

This experiment adopts an extremely \emph{strict} definition of relevance: namely, the only relevant types are the ones that are \emph{explicitly} asserted for the entity in the test partition. Thus, even a super-type (of the true asserted type) would be marked as `irrelevant' if not explicitly asserted itself, since we do not consider transitive closures in the type-only KB. Although a strict measure, it provides a reasonable first benchmark, as it conforms to extensional semantics. 

\begin{figure}[t]
\centering
\includegraphics[height=3.1cm, width=8.7cm]{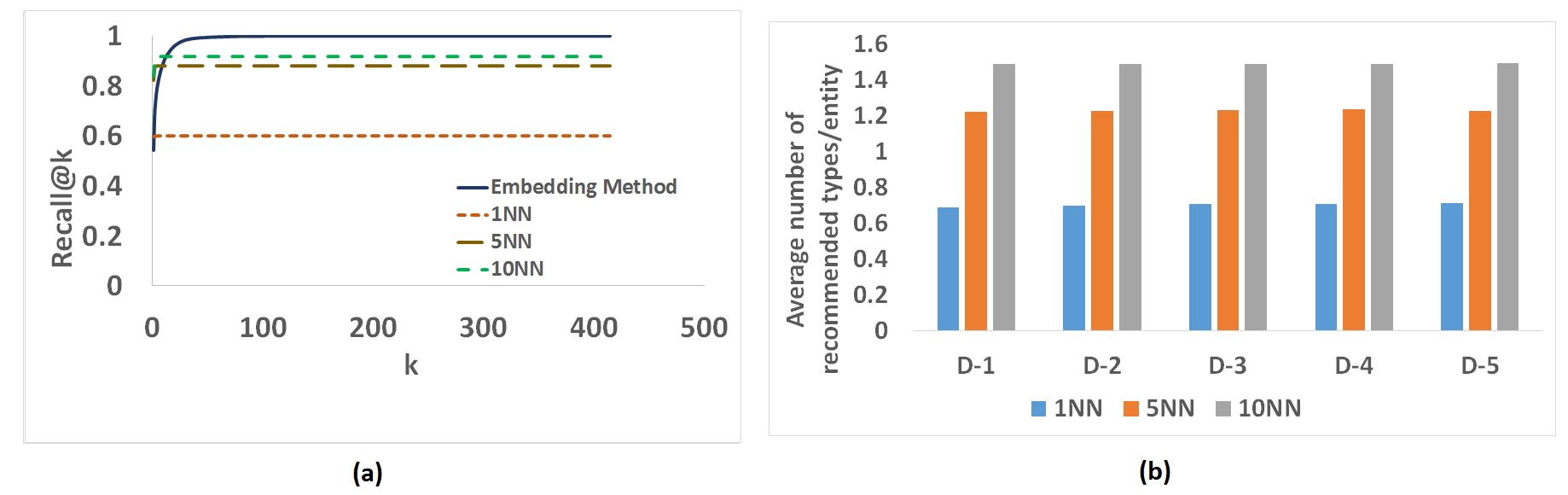}
\caption{Experiment 1 results over all five datasets in Table \ref{datasets}. (a) plots the average Recall@k across all five experimental runs, while (b) illustrates average number of recommended types per entity for each of the datasets and baselines.}
\label{exp-fig1} 
\end{figure}

We evaluate performance using the \emph{Recall@k}\footnote{$k$ in Recall@k should not be confused with $k$ in kNN.} measure from Information Retrieval. Recall@k computes, for each rank $k$ in a ranked list of types, the ratio of true positives to the sum of true positives and false negatives. For each one of the five experimental runs, we compute a single Recall@k measure (for each $k$) by averaging the Recall@k over all 5000 entities. Finally, we compute the average across the experimental runs, and plot the results in
Figure \ref{exp-fig1} (a). We also plotted individual plots for all five experimental runs, which turned out to be qualitatively very similar to Figure \ref{exp-fig1} (a). We omit those figures and a per-run analysis herein due to space constraints.

{\bf Analysis: }Figure \ref{exp-fig1} (a) shows that, even with the strict definition of relevance, although the embedding method starts out with low recall at the highest ranks, it converges with the other methods fairly quickly (between ranks 3-13). Figure \ref{exp-fig1} (b) shows that the baselines return very few non-zero recommendations per entity (fewer than 1.5) and the returned number is \emph{sub-linear} in k: in this respect, the baselines perform `hard' type predictions rather than graded recommendations. In contrast, the embedding method returns a more nuanced probability distribution over the 415 types (per entity), and is more apt for recommendations, as we show in the next experiment.

{\bf Experiment 2: } Although Experiment 1 is appropriate for determining the extensional types of an entity, it takes an overly strict reflection of relevance. Considering the number of triples and unique instances in Table \ref{datasets}, there was usually only one extensional type asserted in the knowledge base. For a better judgment of relevance, we randomly sampled 100 instances from D-1 and pruned the ranked type lists for 10NN (clearly the best performing method in Figure \ref{exp-fig1} (a)) and the embedding method to 10. Because the number of returned types for 10NN was often fewer than 10, we randomly `padded' the rest of the list with DBpedia types, and manually counted the number of topically relevant recommendations in each (10-element) list\footnote{The annotations were performed externally, not by the authors.}. This allows us to compute a single-point Recall@10 score  over the 100 sampled entities. Note that the random padding can only help, not hurt, the Recall@10 score of the baseline. We also asked the annotator to do a side-by-side comparison of the two lists (for each of the 100 entities) and mark the list that is more \emph{topically useful} overall. The annotator was not given the details of the two ranking algorithms; also, all annotations were conducted within a single short time-span.

\begin{table}[t]
\centering
\caption{Top 3 type recommendations for the embedding method and 10NN for five entities from the 100-sample dataset (Experiment 2)}
\begin{tabular}{|p{1.7cm}|p{2.3cm}|p{3.2cm}|} \hline
{\bf Entity}&{\bf Embedding Method Rec.} & {\bf 10NN  Rec.} \\ \hline
{Shenyang\_J-13} & {Aircraft, Weapon, Rocket} & {Aircraft, Weapon, Photographer} \\ \hline
{Amtkeli\_ River} & {River, BodyOfWater, Lake} & {River, Dam, Mountain} \\ \hline
{Melody\_ Calling} & {Album, Single, Band} & {Album, Single, PublicTransitSystem} \\ \hline
{Esau\_ (judge\_royal)} & {Monarch, Loyalty, Noble} & {Noble, Journalist, AdministrativeRegion} \\ \hline
{Angus\_ Deayton} & {Comedian, ComedyGroup, RadioProgram} & {Person, Actor, TelevisionShow} \\ \hline
\end{tabular}
\label{recexamples}
\end{table} 

{\bf Analysis:} The single-point Recall@10 was 0.4712 for the embedding method (averaged across  the 100 manually annotated samples) and 0.1423 for the 10NN with standard deviations of 0.2593 and 0.0805  respectively. Some representative results for both 10NN and the embedding method are presented in Table \ref{recexamples}. The table presents some intuitive evidence that types recommended by the embedding method are more topically relevant than the types recommended by 10NN. The annotator found the embedding method to be topically more useful for 99 of the 100 entities, with 10NN more useful on only a single entity.

{\bf Run-times:} Concerning the empirical run-times, all baseline methods required about 1 hour to process 5000 test instances for the 4-partition training set while the embedding method only required 4 minutes. If the training set is fixed, all methods were found to exhibit linear dependence on the size of the test set. This illustrates why we were forced to sample 5000 test instances (per experimental run) for evaluating kNN, since predicting types on only one of the full five partitions in Table \ref{datasets} would take about 150 hours, which is untenable, even if only strict type predictions (i.e. assertions) are of interest. 

\subsection{Visualizing the Extensional Model}\label{exp2}
Note that our methods never relied on the DBpedia ontology when deriving embeddings and generative model parameters. However, an interesting experiment is to use the intensional semantics of types (e.g. sub-class relationships in the ontology) to visualize the embeddings derived from extensional assertions (the mapping-based type assertion extractions). We perform two visualization experiments using the \emph{unsupervised} t-SNE algorithm \cite{tsne}, a state-of-the-art tool for visualizing high-dimensional points on a 2-D plot.  
\begin{figure}[t]
\centering
\includegraphics[height=3.4cm, width=8.7cm]{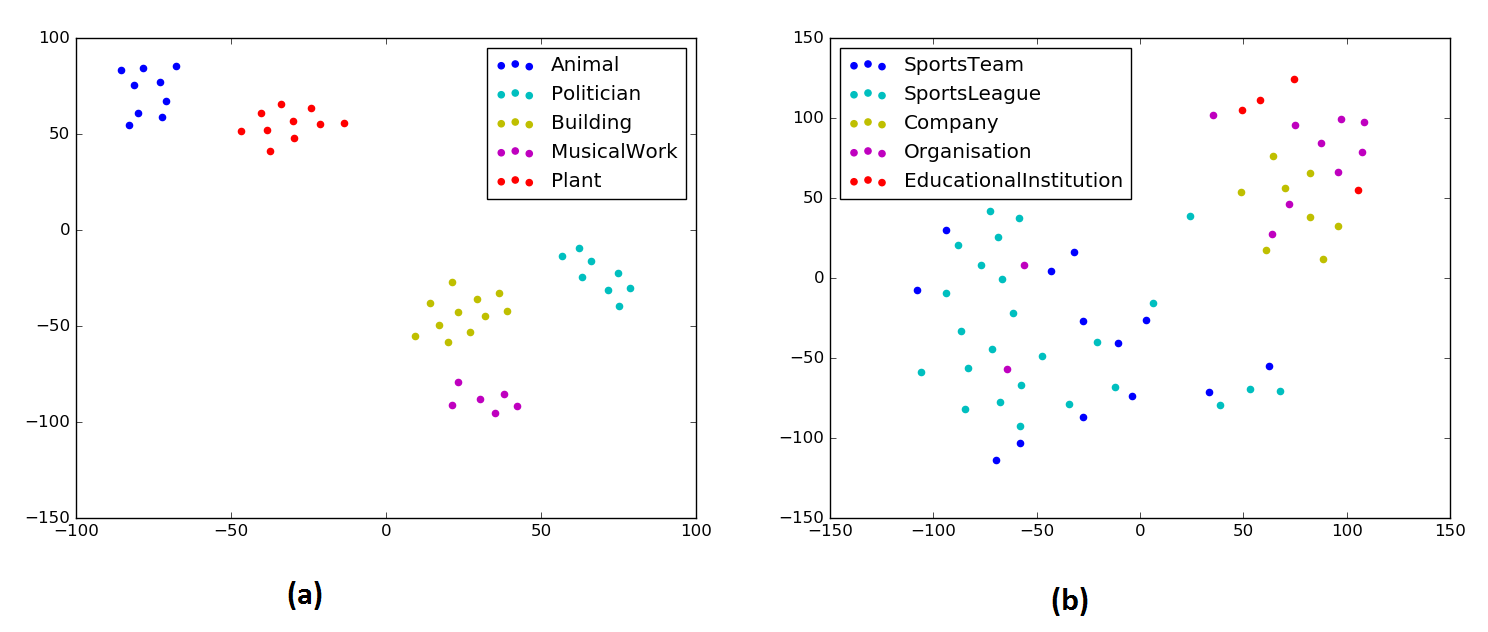}
\caption{Visualizing the extensional type embeddings in terms of the relationships in the domain ontology. The axes do not have intrinsic meaning.}
\label{exp-fig2} 
\end{figure}

{\bf Visualization 1:} We apply t-SNE on a matrix containing the type embedding vectors of all direct sub-types of five sampled types from the DBpedia ontology, namely \emph{Animal, Politican, MusicalWork, Building} and \emph{Plant}. The t-SNE algorithm takes the matrix as input and returns another matrix with the same number of rows but only 2 columns. We plot these points (rows) in Figure \ref{exp-fig2} (a), by assigning labels (i.e. colors) to points\footnote{Because t-SNE is unsupervised, it never accessed the labels during the clustering.} based on their super-type.  The results show five well-separated clusters, with each cluster representing a super-type. In this case, the extensional model is in excellent agreement with the intensional model. Moreover, the clusters also demonstrate other interesting aspects not captured intensionally: e.g. \emph{Building, Politician} and \emph{MusicalWork} (artificial constructs) are much closer to each other, than they are to \emph{Animal} and \emph{Plant} (natural constructs), which form a separate `super' cluster.

{\bf Visualization 2:} We re-run the experiment but over sub-type embeddings of the types \emph{SportsTeam, SportsLeague, Company, Organisation} and \emph{EducationalInstitution}. Note that this set is much more \emph{topically coherent} than the earlier set. The 2D visualization is illustrated in Figure \ref{exp-fig2} (b); the topical coherence (there are now two clusters rather than five) is well-reflected in the figure. The two purple `outliers' on the left cluster are the embeddings for \emph{SportsTeam} and \emph{SportsLeague}, which are themselves sub-types of \emph{Organisation}. 

\subsection{Online Type Clustering of DBpedia}\label{exp3}
The results of Experiment 2 in Section \ref{exp1} illustrated that there are many types in the DBpedia ontology that are clearly related to an entity of interest, even if they are not super-types (or even sub-types) of the entity's type. Given an entity, we ideally want to assign a probability to \emph{each} type. Such a clustering has much potential, including topic modeling of entities and fuzzy reasoning. To achieve a full fuzzy clustering over DBpedia, we used the union of all five datasets in Table \ref{datasets} to compute our type embeddings, and then executed the fuzzy clustering algorithm over all DBpedia entities in the union of all five datasets. The algorithm scaled near-linearly and was able to finish executing over almost 4 million entities and 415 clusters (types) in about 50 hours, collecting over 100 GB of data. We will make this data available on a public server in the near future. Overall, the results (using the extensional type assertions as ground-truth) were found to be highly correlated to the results in Figure \ref{exp-fig1} (a). We are currently in the process of conducting detailed probabilistic analyses on this data, and are also clustering Wikidata embeddings, a much larger dataset than DBpedia.

\section{Conclusion}\label{conclusion}
In this paper, we developed a framework for deriving type embeddings in the same space as a given set of entity embeddings. We devised a scalable data-driven algorithm for inferring the type embeddings, and applied the algorithm to a probabilistic type recommendation task on five DBpedia partitions. Compared to a kNN baseline, the algorithm yields better results on various relevance criteria, and is significantly faster. Visualizations also show that clusters over the type embeddings are intensionally consistent.

{\bf Future Work.} There are many avenues for future work that we have already started exploring. First, we are using the methods in this paper to embed entire ontologies (collections of types and properties) into vector spaces, to enable a combination of distributional and ontological semantic reasoning. Second, we are exploring more applications of embedded types, such as an enhanced version of semantic search, and semantically guided information extraction from structured data. 
Last, but not least, we are conducting broader empirical studies e.g. on datasets other than DBpedia and using knowledge graph embeddings other than the DeepWalk-based RDF2Vec to test the robustness of our type embedding approach to such variations. We are also testing the hypothesis that deriving type embeddings from entity embeddings yields higher quality typing than treating types as part of a knowledge graph and jointly deriving entity and type embeddings. We are also looking to carry out a broader user study than the preliminary study in Experiment 2.   